# *Programming in Logic Without Logic Programming*

Robert Kowalski and Fariba Sadri
*Department of Computing*
*Imperial College London*
*{rak,fs@doc.ic.ac.uk}*

**Abstract**

In previous work, we proposed a logic-based framework in which computation is the execution of actions in an attempt to make reactive rules of the form *if antecedent then consequent* true in a canonical model of a logic program determined by an initial state, sequence of events, and the resulting sequence of subsequent states. In this model-theoretic semantics, reactive rules are the driving force, and logic programs play only a supporting role.

In the canonical model, states, actions and other events are represented with timestamps. But in the operational semantics, for the sake of efficiency, timestamps are omitted and only the current state is maintained. State transitions are performed reactively by executing actions to make the *consequents* of rules true whenever the *antecedents* become true. This operational semantics is sound, but incomplete. It cannot make reactive rules true by preventing their *antecedents* from becoming true, or by proactively making their *consequents* true before their *antecedents* become true.

In this paper, we characterize the notion of reactive model, and prove that the operational semantics can generate all and only such models. In order to focus on the main issues, we omit the logic programming component of the framework.

## 1  Introduction

State transition systems play an important role in many areas of Computing. They underpin the operational semantics of imperative programming languages, the dynamic behavior of database management systems, and many aspects of knowledge representation in artificial intelligence. In many of these systems, state transitions are performed by executing reactive rules of the form *if antecedent then consequent*, which describe relationships between earlier and later states and events. Such reactive rules occur explicitly as condition-action rules in production systems, event-condition-action rules in active databases, and transition rules in Abstract State Machines (Gurevich 2000). They are implicit in Statecharts (Harel 1987) and BDI agents plans (Rao and Georgeff 1995). They are the core of Reaction RuleML (Paschke et al. 2012).

Despite the apparently logical *syntax* of reactive rules in these systems, hardly any of these systems give *if-then rules* a logical interpretation. In this paper, we investigate the semantics of a logical language, KELPS, in which programs are sets of reactive rules of the form $\forall X \, [antecedent \rightarrow \exists Y \, [consequent]]$ in classical, first-order logic (FOL), and computation is understood as generating a sequence of state transitions with the purpose of making the reactive rules true.

KELPS (Kowalski and Sadri 2012) is the reactive Kernel of LPS (Kowalski and Sadri 2009, 2010, 2011, 2012, 2014, 2015), a Logic-based agent and Production System language, which combines reactive rules and logic programs. KELPS is obtained from LPS by dropping the logic programming component of LPS. It is in this sense that KELPS is a language for programming in logic without logic programming.

The operational semantics (OS) of KELPS is similar to that of imperative reactive rule languages, which maintain only a single current state, using destructive state transitions. However the model-theoretic semantics of KELPS combines all the states into a single model, by associating time stamps with facts, actions and external events.

In (Kowalski and Sadri 2010, 2011, 2014, 2015), we showed that the OS of LPS (and therefore of KELPS) is *sound*: Any sequence of states and events that the OS recognizes as solving the computational task generates a model that makes the reactive rules *true*. In this paper, we investigate the completeness of the OS of KELPS, and show that the OS can generate all *reactive models*, in which the consequents of reactive rules are made *true* after their antecedents become *true*. However, the OS of KELPS (and therefore of LPS) is *incomplete,* because it can generate *only* reactive models.

The OS of KELPS cannot generate models that *proactively* make consequents *true* whether or not their antecedents become *true*; that *preventively* make antecedents *false* to avoid making their consequents *true*; or that make their antecedents *true*, and are then forced to make their consequents *true*. Moreover, it does not generate models that contain actions that are *irrelevant* to the computational task.

Because the OS of KELPS is similar to that of imperative reactive rule languages, the incompleteness of the KELPS OS shows that the operational semantics of conventional reactive system languages are also incomplete, if their reactive rules are read as logical implications.

LPS (and therefore KELPS) is a scaled-down and optimised version of abductive logic programming (ALP) (Kakas et al. 1998). There exist proof procedures for ALP that can generate proactive and preventative models, but avoid generating obviously irrelevant actions. In section 6.2, we discuss the relationship between KELPS/LPS and ALP.

Figure 1 illustrates KELPS and some of the different kinds of models that are allowed by the model-theoretic semantics of KELPs. There is a single reactive rule stating that if you see a wolf at time $T$ then you cry wolf at time $T+1$. An external event, *see-wolf*, occurs at time 3. The reactive model includes all and only the external events and the actions (here, the single action *cry-wolf* at time *4*) that are motivated by the reactive rule and triggered by the external events.

| | |
|---|---|
| Reactive rule: | $\forall T$ [*see-wolf(T)* $\rightarrow$ *cry-wolf(T+1)*] |
| External event: | *see-wolf(3)* |
| Reactive model: | *see-wolf(3), cry-wolf(4)* |
| Proactive model: | *cry-wolf(1), cry-wolf(2), see-wolf(3), cry-wolf(4)* |
| Model with irrelevant action: | *see-wolf(3), cry-wolf(4), drink(4)* |

Fig. 1: Example Models of Reactive Rules in KELPS.

The example in Figure 2 is a variant of that in Figure 1. The reactive rule in Figure 2 states that you cry wolf if you see a wolf and you are outside. The state records whether or not the agent is outside. The causal theory updates the state as a result of both external events and actions. The agent is initially outdoors. Both models in Figure 2 include all the timestamped facts belonging to any state, all the timestamped external events, and all the timestamped actions motivated by the reactive rule. The non-reactive, *preventative* model includes an action of going inside, which prevents the need for crying wolf. Of course, another non-reactive model can include both actions.

| | |
|---|---|
| Reactive rule: | $\forall T$ [*see-wolf(T)* ∧ *outdoors(T)*→ *cry-wolf(T+1)*] |
| Causal Theory: | *terminates(go-inside, outdoors)* |
| | *initiates(go-outside, outdoors)* |
| Initial state $S_0$ at time *0*: | *outdoors* |
| External event: | *see-wolf(3)* |
| Reactive model: | *outdoors(0), outdoors(1), outdoors(2), outdoors(3), ....,* |
| | *see-wolf(3), cry-wolf(4)* |
| Preventative model: | *outdoors(0), outdoors(1), go-inside(1), see-wolf(3)* |

Fig. 2: Example Models of Reactive Rules in KELPS.

In this paper, we characterize the reactive models *I* generated by the KELPS OS. These models all have the property that every action in *I* is motivated by being an instance of an action that occurs explicitly in the consequent of a rule whose earlier conditions (in the antecedent or consequent of the rule) are already *true*.

In the remainder of the paper, we present KELPS, its model-theoretic and operational semantics, the relationship between the two semantics, the relationship with related work, and future work.

This paper extends an earlier paper (Kowalski and Sadri 2014) by including proofs of all the theorems (in the Appendix), extending KELPS to allow more general FOL conditions, simplifying many of the definitions, and including an extensive comparison with related approaches.

## 2  KELPS

The example in Figure 3 illustrates additional features of the language. In this example, the consequent of the rule consists of two alternative plans with deadlines: When an order is received from a reliable customer, then the item needs to be dispatched, and an invoice needs to be sent within 3 time units of receiving the order. Alternatively, an apology needs to be sent within 5 time units. Temporal constraints are defined by an auxiliary theory, which is not presented here. They are solved by means of a constraint solver, as in constraint logic programming (Jaffar and Lassez 1987).

Actions in KELPS can be executed concurrently. Consistency is maintained by monitoring the preconditions of sets of events. In this example, a precondition ensures that the same item is not dispatched to two different customers at the same time. We will see later that preconditions can also be used to prevent sending an apology if the item has already been dispatched.

The semantics of KELPS is non-deterministic: It does not matter which alternative plan is chosen. However in practice, some alternatives are better than others. For example, it may be a good strategy to try first a plan with the earliest deadline, and to try later an alternative plan with a later deadline if the earlier plan has failed. Any actions performed in the earlier, partially executed plan are committed choices, which cannot be undone by rolling back time. At best, their effects can be undone only by performing compensating actions later in time.

In a practical implementation, for most applications, it would be desirable to provide a method to control the choice of alternatives. However, for lack of space, we do not address such control issues in this paper.

| | |
|---|---|
| Reactive rule: | $\forall C, Item, T1\ [orders(C, Item, T1) \land reliable(C, T1)$ <br> $\rightarrow \exists\ T2, T3\ [dispatch(C, Item, T2) \land send\text{-}invoice(C, Item, T3) \land$ <br> $T1 < T2 \leq T3 \leq T1 + 3]$ <br> $\lor\ [send\text{-}apology(C, Item, T4) \land T1 < T4 \leq T1 + 5]]]$ |
| Causal theory: | $initiates(send\text{-}invoice(C, Item), payment\text{-}due(C, Item))$ <br> $terminates(pays\text{-}invoice(C, Item), payment\text{-}due(C, Item))$ <br> $dispatch(C1, Item, T) \land dispatch(C2, Item, T) \land C1 \neq C2 \rightarrow false$ |
| Initial state: | $reliable(bob)$ |

Fig. 3: An Example with a Complex Consequent.

The operational semantics of KELPS maintains a single current state $S_i$ at time $i$. It reasons with the reactive rules, to generate a set of actions $acts_{i+1}$, which it combines with a set of external events $ext_{i+1}$, to produce a consistent set of concurrent events $ev_{i+1} = ext_{i+1} \cup acts_{i+1}$. The events $ev_{i+1}$ are used to update the current state $S_i$, generating the successor state $S_{i+1} = succ(S_i, ev_{i+1})$ by deleting facts that are terminated by the events $ev_{i+1}$ and adding facts that are initiated by $ev_{i+1}$.

In KELPS, states are represented by sets of atomic sentences (also called *ground atoms*, *facts* or *fluents*). Events are also represented by atomic sentences. Such sets of atomic sentences can be understood either syntactically as theories or sematically as model-theoretic structures. It is this second, model-theoretic interpretation that underpins the logical semantics of KELPS.

States and events can be represented with or without timestamps. The representation without timestamps (e.g. *outdoors, reliable(bob)*) facilitates destructive updates, because if a fact is not terminated by a set of events then the fact without timestamps simply persists from one state to the next. However, the representation with timestamps (e.g. *outdoors(0), reliable(bob, 0)*) makes it possible to combine all states and events into a single model-theoretic structure.

### 2.1  *Vocabulary*

KELPS is a first-order, sorted language, including a sort for time. In the version of KELPS presented in this paper, we assume that time is linear and discrete, and that the succession of timepoints is represented by the ticks of a logical clock, where 1, 2, ... stand for $s(0), s(s(0)), ...., t+1$ stands for $s(t)$ and $t+n$ stands for $s^n(t)$. Thus $S_i$ represents the state at time $i$, and $ev_{i+1}$ represents the set of events taking place in the transition from state $S_i$ to $S_{i+1}$. Other representations of time are also possible.

**Predicates:** The predicate symbols of the language are partitioned into sets representing fluents, events, auxiliary predicates and meta-predicates:

*Fluent* predicates represent facts in the states $S_i$. The last argument $i$ of a timestamped fluent atom $p(t_1, ..., t_n, i)$ is a time parameter, representing the time $i \geq 0$ of the state $S_i$ to which the fluent belongs. The unstamped fluent atom $p(t_1, ..., t_n)$ is the same atom without this timestamp.

*Event* predicates represent events contributing to the transition from one state to the next. The last argument of a timestamped event atom $e(t_1, ..., t_n, i)$ is a time parameter, representing the time $i \geq 1$ of the successor state $S_i$. The unstamped event atom $e(t_1, ..., t_n)$ is the same atom without this time parameter. Event predicates are partitioned into *external event predicates* and *action predicates*.

Fluent and event predicates can have time parameters, called *reference times*, that are not timestamps. For example in *deadline(Task, D, T)*, the time parameter $D$ is a reference time, which expresses that at time $T$ the deadline for *Task* is $D$.

*Auxiliary* predicates are of two kinds: *Time-independent predicates*, for example *isa(book, item)*, do not include time parameters. *Temporal constraint predicates*, including inequalities of the form $T1 < T2$ and $T1 \leq T2$ between timepoints, and functional relationships among timepoints, such as *max(T1, T2, T)* and *min(T1, T2, T)* have only time parameters.

In KELPS, temporal constraints constrain the timestamps of fluents and events. As a consequence, every temporal constraint in a reactive rule contains at least one time parameter that occurs as a timestamp in a fluent or event atom of the rule.

In LPS, auxiliary predicates are defined by logic programs. In KELPS, they are defined more simply by a (possibly infinite) set ***Aux*** of atomic sentences. In the case of auxiliary temporal constraint predicates, this assumption is equivalent to the assumption made in the semantics of constraint logic programming (CLP). The KELPS OS exploits this relationship with CLP by using a constraint solver to simplify temporal constraints and to check them for satisfiability.

The *meta-predicates* consist of the two predicates *initiates(events, fluent)* and *terminates(events, fluent)*, which specify the post-conditions of events and perform state transitions, as illustrated in Figure 2. The first argument is a set of events, to cater for the case where two events together have different effects from the individual events on their own (as when you buy two books and get the cheaper one for half price; or when two people push a heavy object that cannot be moved by one person alone). The second argument is a fluent without timestamps. In LPS, these meta-predicates are defined by logic programs. In KELPS, they are defined by atomic sentences[1] in a *causal theory* ***C***, which also contains constraints on the preconditions and co-occurrence of events.[2]

**Notation:** If $S_i$ is a set of fluents without timestamps, representing a single state, then $S_i^*$ represents the same set of fluents with the same timestamp $i$. If *events$_i$* is a set of events without timestamps, all taking place in the transition from state $S_{i-1}$ to state $S_i$, then *events$_i^*$* represents the same set of events with the same timestamp $i$.

---

[1] In the examples, in Figure 3 and elsewhere in the paper, we use variables in the definitions of auxiliary predicates as a shorthand for the set of all the well-sorted ground instances of the definitions.

[2] In some earlier papers, this causal theory was called a "domain theory".

## 2.2 KELPS Framework

***Definition.*** A KELPS framework (or program) is a triple <***R***, ***Aux***, ***C***>, where ***R*** is a set of reactive rules, ***Aux*** is a set of ground atoms defining auxiliary predicates, and ***C*** is a causal theory.

Rules in ***R*** are constructed from formulas that represent complex patterns of states and events, expressed as conjunctions of *FOL conditions* and temporal constraints. Operationally, an *FOL* condition is a query to ***Aux*** ∪ $S_i^*$ ∪ $ev_i^*$, which is the timestamped state at time *i*, augmented with the most recent set of events and the definition of the time-independent auxiliary predicates. For example, the *FOL* condition:

$$\forall Item \; \forall D \; [manages(M, D, T) \land item(Item, D) \rightarrow instock(Item, T)]$$

behaves as a query that returns managers *M* all of whose departments *D* have all of their items *Item* in stock at time *T*. The variables *Item* and *D* are said to be bound in the condition, and the variables *M* and *T* are unbound in the condition. We will see later that, depending where the condition occurs in a reactive rule, the variables *M* and *T* will be either universally or existential quantified in the rule.

***Definition.*** An *FOL condition* is an FOL formula containing exactly one timestamp, which is either a constant or a variable, whose atoms are either fluent atoms, event atoms or time-independent auxiliary predicates. If the timestamp is a variable, then it is not bound by any quantifier in the FOL condition.

***Definition.*** A *complex pattern of states and events* (*complex* for short) is a possibly empty conjunction of FOL conditions and temporal constraints. All time variables in temporal constraints in the complex occur as time parameters in FOL conditions of the complex or are functionally dependent on such time parameters.

For example, *T3* is functionally dependent on *T1* and *T2 in max(T1, T2, T3)*, and *T2* is functionally dependent on *T1* in *plus(T1, 3, T2)*.

***Definition.*** A *reactive rule* (or simply *rule*) is a sentence of the form:
 ∀*X* [*antecedent* → ∃*Y* [*consequent*]] where:
- *X* is the set of all variables, including time variables, occurring in *antecedent* and not bound in FOL conditions. *Y* is the set of all variables, including time variables, occurring only in *consequent* and not bound in FOL conditions.
- *consequent* is a disjunction *consequent₁* ∨ ...∨ *consequentₙ*.
- *antecedent* and each *antecedent* ∧ *consequentᵢ* is a complex.
- For every substitution σ that replaces the time variables in *X* and *Y* by ground times and such that the temporal constraints in *antecedent* σ and *consequent* σ are *true* in ***Aux***, all timestamps in FOL conditions in *consequent* σ are later than or equal to all timestamps in FOL conditions in *antecedent* σ.

- Every temporal constraint in $consequent_i$ contains at least one timestamp variable that occurs in an FOL condition in $consequent_i$ or is functionally dependent on such timestamp variables.

Intuitively, the next-to-last bullet restricts reactive rules to ones whose *antecedent* is a conjunction of FOL conditions about the past or present and whose *consequent* is a disjunction of conjunctions of FOL conditions about the present or future. The last bullet prevents such rules as $p(T1) \rightarrow q(T2) \land T1<10 \land T1<T2$, with a constraint $T1<10$ in the *consequent* that only constrains a timestamp in the *antecedent*.

Because of the restrictions on quantifiers, and because of the logical equivalence $\exists Y\, [p \lor q] \Leftrightarrow \exists Y\, p \lor \exists Z\, q$, we can omit the quantifiers $\forall X$ and $\exists Y$, and simply write *antecedent* $\rightarrow$ *consequent* or *antecedent* $\rightarrow$ *consequent$_1$* $\lor ... \lor$ *consequent$_n$*.

Variables that are unbound in an FOL condition become bound either universally or existentially, depending on where the condition occurs in a rule. For example, if the condition $\forall Item\, \forall D\, [manages(M, D, T) \land item(Item, D) \rightarrow instock(Item, T)]$ occurs in the antecedent of a rule, then $M$ and $T$ are bound by the universal quantifiers of the rule. If the condition occurs in the consequent of a rule, and $M$ and $T$ do not occur in the antecedent of the rule, then $M$ and $T$ are bound by the existential quantifiers of the consequent of the rule. To avoid ambiguity, if an FOL condition is the *consequent* of a rule with an empty antecedent, then we write the rule in the form *true* $\rightarrow$ *consequent*.

Note that, in the operational semantics, all the components of an FOL condition are evaluated together in $\mathbf{Aux} \cup S_i^* \cup ev_i^*$. Several such FOL conditions can be evaluated at the same time $i$, if their timestamps can all be unified to time $i$. Note also that, if *antecedent* is empty, then *antecedent* is equivalent to *true*. If *consequent* is empty, then *consequent* is equivalent to *false*.

An FOL condition that contains an action atom can be evaluated in $\mathbf{Aux} \cup S_i^* \cup ev_i^*$, like any other FOL condition. However, an action atom *act* that is a conjunct of an FOL condition of the form[3] *conjunct1* $\land$ *act* $\land$ *conjunct2* in the consequent of a rule can also be selected as a candidate for execution in the transition to the next state $S_{i+1}$. Such action atoms are called *bare action atoms.*

**Definition.** Let *consequent$_i$* of a rule be of the form *conjunct1* $\land$ *act* $\land$ *conjunct2* where *act* is an action atom, then *act* is a *bare action atom* in *consequent$_i$*.

State transitions are performed by means of a causal theory, which imposes preconditions on sets of events and defines the postconditions of sets of events.

**Definition.** A *causal theory*, $C = C_{post} \cup C_{pre}$, consists of two parts: $C_{post}$ is a set of atomic sentences defining the predicates *initiates* and *terminates*. $C_{pre}$ is a set of sentences of the form *current(T-1)* $\land$ *events(T)* $\rightarrow$ *false,* where *current(T-1)* is a (possibly empty) FOL condition with timestamp *T-1*, *events(T)* is a non-empty FOL condition with timestamp *T* containing no fluents, and all variables not explicitly bound in FOL conditions are implicitly universally quantified.

---

[3] For simplicity we say that a formula has a particular form when we mean that the formula can be rewritten into that form simply by reordering conjunctions and disjunctions, taking commutativity and associativity into account.

The syntax of $C_{pre}$ allows the specification of such typical preconditions for the execution of a single action as:

$\neg$ *in-stock(Item, T-1)* $\wedge$ *dispatch(C, Item, T)* $\rightarrow$ *false*

where *in-stock(Item)* is a fluent, initiated and terminated by such actions as *stock(Item)* and *dispatch(C, Item)*, respectively. The syntax of $C_{pre}$ also allows the prohibition of such concurrent sets of events as:

*dispatch(C1, Item, T)* $\wedge$ *dispatch(C2, Item, T)* $\wedge$ *C1* $\neq$ *C2* $\rightarrow$ *false*

It is also possible to specify that certain actions must co-occur:

*leave-house(T)* $\wedge$ $\neg$ *take-keys (T)* $\rightarrow$ *false.*

In the example of Figure 3, sending an apology if an item has already been dispatched or is being dispatched can be prevented by means of the preconditions:

*dispatched(C, Item, T)* $\wedge$ *send-apology(C, Item, T)* $\rightarrow$ *false*
*dispatch(C, Item, T)]* $\wedge$ *send-apology(C, Item, T)* $\rightarrow$ *false*

where *dispatched(C, Item)* is a fluent initiated by the action *dispatch(C, Item)*.

The use of atomic sentences to define the predicates *initiates* and *terminates* is similar to the use of add-lists and delete-lists in STRIPS (Fikes and Nilsson 1972). However, it is more general, because the first argument of both predicates is a *set* of events. Defining the fluents initiated and terminated by sets of concurrent events explicitly by means of atomic sentences is not very practical, but it clarifies the model-theoretic semantics and simplifies the operational semantics. Moreover, it paves the way for the more practical representation in which *initiates* and *terminates* are defined by logic programs in LPS.

### 3 The KELPS Model-theoretic Semantics

In the model-theoretic semantics of KELPS, the truth values of the rules **R** and the preconditions $C_{pre}$ are defined according to the standard, non-modal semantics of classical first-order logic. This contrasts with the semantics of modal logics, in which states are represented by possible worlds, linked by accessibility relations.

In the operational semantics of KELPS, states are updated by adding and deleting fluents. Fluents that are not affected by the update are left untouched:

**Definition.** If <**R**, **Aux**, **C**> is a KELPS framework, *S* is a set of unstamped fluents, representing a single state, and *ev* is a set of unstamped events, representing concurrent events, then the associated *successor state* is:

*succ*(*S*, *ev*) = (*S* − {*p* | *terminates(ev, p)* $\in$ $C_{post}$ }) $\cup$ {*p* | *initiates(ev, p)* $\in$ $C_{post}$}.

In the model-theoretic semantics, fluents and events are all timestamped, so they can be included in a single model-theoretic structure $M = Aux \cup S^* \cup ev^*$:

**Notation.** If $S_0$ is an initial state, $ext_1, \ldots, ext_i, \ldots$, is a sequence of sets of external events and $acts_1, \ldots, acts_i, \ldots$ is a sequence of sets of actions, then:

$S^* = S_0^* \cup \ldots \cup S_i^* \cup \ldots$ where $S_{i+1} = succ(S_i, ev_{i+1})$
$ev^* = ev_0^* \cup \ldots \cup ev_i^* \cup \ldots$ where $ev_0 = \{\}$ and $ev_i = ext_i \cup acts_i$, for $i \geq 1$.

Computation in a conventional reactive system consists in generating a stream $act_1, \ldots, act_i, \ldots$ of actions in response to a stream $ev_1, \ldots ev_i, \ldots$ of external events and previously generated actions. Computation in KELPS is similar, but it has a purpose, namely to make the reactive rules and the preconditions of actions *true*:

***Definition.*** Given a KELPS framework $\langle R, Aux, C \rangle$ and initial state $S_0$, the *computational task* is, for every $i \geq 0$, and for every sequence $ev_0 \ldots ev_i$ of sets of external events and previously generated actions, to generate a set $acts_{i+1}$ of actions such that $R \cup C_{pre}$ is *true* in the Herbrand interpretation $M = Aux \cup S^* \cup ev^*$.

The definition of truth for reactive rules is the classic definition for sentences of FOL. As a consequence, the computational task allows the generation of actions that make the rules *true* by making their antecedents *false*, or by making their consequents *true* whether their antecedents are *true* or *false*. It also allows the performance of actions that are irrelevant to the task. These kinds of "preventative", "proactive" or "irrelevant" actions cannot be generated by conventional reactive systems. Nor can they be generated by KELPS. In this paper, we identify the kind of reactive models that are generated by the KELPS operational semantics.

Note that in KELPS the generated actions $acts_{i+1}$ need not be a direct reaction to the current situation $S_i^* \cup ev_i^*$. They can be a partial response to earlier situations.

### 3.1 Herbrand interpretations

The semantics of Herbrand interpretations is a simplified version of the standard semantics of first-order logic.

***Definition.*** Given a sorted first-order language, the *Herbrand universe U* is the set of all well-sorted *ground* (i.e variable-free) terms that can be constructed from the non-empty set of constants and function symbols of the vocabulary. The *Herbrand base* is the set of all well-sorted ground atoms that can be constructed from the predicate symbols and the ground terms of the vocabulary. A *Herbrand interpretation* is a subset of the Herbrand base. A *Herbrand model M* of a set *S* of sentences is a Herbrand interpretation such that every sentence in *S* is *true* in *M*.

One difference from the standard definition of truth is the base case: If $I$ is a Herbrand interpretation, then a ground atom $A$ is *true* in $I$ if and only if $A \in I$. The other difference is the definition of truth for universally and existentially quantified sentences: A sentence of the form $\forall X\ s(X)$ is true if and only if for all $t \in U$, where $t$ has the same sort as *X*, the sentence *s(t)* is true. Similarly, ∃*X s(X)* is true if and only if for some $t \in U$, where *t* has the same sort as *X*, the sentence *s(t)* is true. For this to be sensible, the Herbrand universe *U* needs to be non-empty, as in the standard definition of truth.

Thus, a rule ∀*X* [*antecedent*→ ∃*Y* [*consequent₁* ∨ ...∨ *consequentₙ*]] is *true* in *I* if and only if, for every ground instance *antecedent* σ that is *true* in *I*, there exists a ground instance *consequentᵢ* σ θ that is also *true* in *I*. Here the substitutions σ and θ replace the variables *X* and *Y*, respectively, by terms of the appropriate sort in the Herbrand universe *U*. For simplicity, we assume that, except for time parameters, all fluents have the same ground instances over *U* in all states.

### *3.2 The temporal structure of KELPS interpretations*

The timestamping of fluents and events, and the restrictions on the syntax of KELPS provide Herbrand interpretations of KELPS programs with a rich structure of sub-interpretations. This structure is captured by the following theorem, which is an immediate consequence of the definition of truth.

***Theorem 1.*** Given a KELPS framework <***R***, ***Aux***, ***C***>, initial state $S_0$, and sequence of sets of events $ev_0 \ldots ev_i$:
1. If *s* is a conjunction of temporal constraints whose time parameters are all ground, then *s* is *true* in ***Aux*** ∪ ***S*** ∪ ***ev**** if and only if *s* is *true* in ***Aux***.
2. If *s* is a conjunction of FOL conditions and temporal constraints whose time parameters are all ground, then:
   a. If all the timestamps in *s* are the same time *i*,
      then *s* is *true* in ***Aux*** ∪ ***S**** ∪ ***ev**** if and only if *s* is *true* in ***Aux*** ∪ $S_i$* ∪ $ev_i$*.
   b. If *i* is the latest timestamp in *s*, then *s* is *true* in ***Aux*** ∪ ***S**** ∪ ***ev****
      if and only if *s* is *true* in ***Aux*** ∪ $S_0$* ∪ ... $S_i$* ∪ $ev_0$* ∪ ... ∪ $ev_i$*.

There is an obvious similarity with the possible world semantics of modal logic. Each ***Aux*** ∪ $S_i$* ∪ $ev_i$* is similar to a possible world, and the single interpretation ***Aux*** ∪ ***S**** ∪ ***ev**** is similar to a complete frame of possible worlds and accessibility relations. In the possible world semantics, fluents belong to possible worlds, and events belong to accessibility relations. But in KELPS, all fluents and events are timestamped and contained in the single interpretation ***M*** = ***Aux*** ∪ ***S**** ∪ ***ev****.

### *3.3 Sequencing*

The temporal constraints of a complex impose a partial order on the timestamps of the FOL conditions in the complex. Although these timestamps are partially ordered, the complex is used to recognize or generate linearly ordered sequences of states satisfying the FOL conditions of the complex.

It is useful to have a notation in the meta-language that distinguishes between the different sequences represented by the same complex. This notation is not part of the KELPS object language, but is useful for defining reactive interpretations and the operational semantics of KELPS. Intuitively, a sequencing of the form *earlier* < *later* means that the FOL conditions in *earlier* can be evaluated (recognised or generated) before the FOL conditions in *later*.

***Definition.*** Let *earlier* and *earlier* ∧ *later* be complexes.[4] Then *earlier* ∧ *later* has a *sequencing* of the form *earlier* < *later* (or of the form *earlier* ≤ *later*) if and only if there exists a substitution σ for all the time variables in *earlier* ∧ *later* such that:
- all the temporal constraints in *earlier* σ ∧ *later* σ are true in ***Aux***
- all the timestamps in FOL conditions in *earlier* σ are earlier than (or earlier than or equal to) all the timestamps in FOL conditions in *later* σ.

Notice that both $p(T) < true$ and $true < p(T)$ are allowed sequencings of $p(T)$. For example, the complex $p(T1) \land q(T2)$ has the strict sequencings: $p(T1) < q(T2)$, $q(T2) < p(T1)$, $true < p(T1) \land q(T2)$ and $p(T1) \land q(T2) < true$.

In some of our earlier papers, we allowed actions selected for execution to contain non-timestamp variables. These variables are instantiated, when they are successfully executed, as feedback from the environment. In this paper, we restrict the selection of actions to ones that have no such variables. For this purpose, we require KELPS frameworks to be range-restricted:

***Definition.*** A KELPS framework <***R***, ***Aux***, ***C***> is *range restricted* if and only if, for every bare action atom *act* containing non-timestamp variables in a rule in ***R*** of the form *antecedent* → [*other* ∨ [*earlier* ∧ *act* ∧ *rest*]], there is a sequencing *antecedent* ∧ *earlier* < *act* ∧ *rest* such that all the non-timestamp variables in *act* occur in *antecedent* or *earlier*.

### 3.4 Reactive interpretations

Figure 1 and Figure 2 exemplify different kinds of models of a KELPS program. The following definition characterizes reactive interpretations and models. Loosely speaking, an action occurs in a reactive interpretation if and only if it occurs as a bare action atom in one of the alternative consequents of an instance of a reactive rule, and all earlier FOL conditions in the antecedent and the alternative consequent of the instance of the rule are already *true* in the interpretation before the time of the action.

***Definition.*** Given a range restricted KELPS framework <***R***, ***Aux***, ***C***>, initial state $S_0$ and set ***ev*** of timestamped events, let $C_{pre}$ be *true* in ***I*** = ***Aux*** ∪ ***S**** ∪ ***ev****, and let ***ev**** = ***ext**** ∪ ***acts**** be a partitioning of ***ev**** into external events ***ext**** and actions ***acts****. Then ***I*** is *reactive* if and only if, for every action *action* ∈ ***I***, there exists a rule $r \in ***R***$ of the form *antecedent* → [*other* ∨ [*earlier* ∧ *act* ∧ *rest*]], and there exists a substitution σ such that *r* σ *supports action*, in the sense that:
  a)   *action* is *act* σ
  b)   *antecedent* σ   ∧ *earlier* σ  < *act* σ ∧ *rest* σ
  c)   *antecedent* σ   ∧ *earlier* σ  ∧ *act* σ  is *true* in ***I***.
***I*** is a *reactive model* of <***R***, ***Aux***, ***C***> if and only if ***I*** is a reactive interpretation and ***R*** is *true* in ***I***.

---

[4] It is not sufficient to require *later* to be a complex because we want to allow *later* to have temporal constraints with time variables occurring in *earlier*.

Note that condition (b) allows *rest* σ to be false in ***I***.

## 4 The KELPS Operational Semantics

The operational semantics exploits the internal structure of KELPS interpretations ***Aux*** ∪ ***S**** ∪ ***ev**** to generate them by *progressively* extending a partial interpretation ***Aux*** ∪ $S_0^*$ ∪ ... $S_i^*$ ∪ $ev_0^*$ ∪ ... $ev_i^*$ one step at a time. Moreover, it does so by maintaining only the unstamped current state $S_i$ and the events $ev_i$ that gave rise to $S_i$, without remembering earlier states and events.

To recognise complex sequences of states and events in the *antecedents* of rules without remembering past states and events, the OS maintains a current set of partially evaluated rules $R_i$, which need to be monitored in the future. For example, suppose that $R_i$ that contains the rule:

*cry-wolf(T)* ∧ ¬*help-arrives(T+1)* → *cry-wolf(T+2)*

and that $ev_i^*$ contains the event occurrence *cry-wolf(i)*. Then the OS evaluates the condition *cry-wolf(T)* in the augmented current state and adds to $R_i$ the new rule:

¬*help-arrives(i+1)* → *cry-wolf(i+2)*

The OS also maintains a goal state $G_i$ containing partially evaluated alternative plans to be made *true* in the future. For example, if $ev_{i+1}^*$ does not contain the event occurrence *help-arrives(i+1)*, then the OS evaluates ¬*help-arrives(i+1)* in the augmented current state to *true*, and adds to $G_{i+1}$ the new, top-level goal *cry-wolf(i+2)*.

Logically, a goal state $G_i$ is a conjunction $G_{i1}$ ∧ ... ∧ $G_{in}$, where each $G_{ij}$ is a disjunction of partially evaluated alternative plans for making true the consequent of an instance of a rule whose antecedent has already become true. To be more precise, each disjunct in $G_{ij}$ is the instantiated remainder *later* σ of a rule *antecedent* → [*other* ∨ [*earlier* ∧ *later*]] in ***R*** whose earlier part *antecedent* σ ∧ *earlier* σ is already *true* in the partial interpretation ***Aux*** ∪ $S_0^*$ ∪ ... $S_i^*$ ∪ $ev_0^*$ ∪ ... $ev_i^*$ generated so far. Because of their similarity to goal clauses in logic programming, such disjuncts *later* σ are also called *goal clauses* in KELPS.

Operationally, the goal state $G_i$ is a set (conjunction) of independent threads $G_{ij}$, and each thread is a goal tree. The root node is the instantiated consequent of a rule whose antecedent has already become true. The non-root nodes are goal clauses. The goal tree representation helps to structure the search space of alternative plans, and to guide the search for alternatives. If the goal trees are searched in a depth-first fashion, then they can be implemented by stacks, as in Prolog. Backtracking is possible, but previously generated actions and states cannot be undone.

The following abstract specification of the OS ignores many optimizations that can improve efficiency. These are described in (Kowalski and Sadri 2010, 2011, 2012, 2014). Some of these optimizations restrict the models that can be generated, and hence affect the relationship between the interpretations generated by the OS and the interpretations sanctioned by the definition of reactive interpretation.

In the following definition, the OS is presented as an agent cycle. At the end of each cycle, external events are input and combined with selected actions. The result-ing combined set of events is used to update the current state. In other versions of the OS, these updates were performed at the beginning of the cycle.

***Definition. The OS Cycle.*** Given a range restricted KELPS framework <*R*, *Aux*, *C*> and an initial state $S_0$, let $ev_0 = \{\}$. Let $G_0$ be the goal state obtained by creating, for every rule *r* in *R* of the form *true* → *consequent*, a goal tree with *consequent* at the root, and adding each disjunct of *consequent* whose constraints are satisfiable in *Aux* as a child of the root. Let $R_0$ be *R* without these rules.

For $i \geq 0$, given $S_i$, $R_i$, $G_i$ and $ev_i$, the *i-th* cycle consists of the following steps:

**Step 1. Evaluate antecedents**. (a) For every sequencing *current* $\theta$ < *later* $\theta$ of the antecedent of an instance $r\theta$ of a rule *r* of the form *current* ∧ *later* → *consequent* in $R_i$, where *current* is a non-empty complex, add *later* $\theta$ → *consequent* $\theta$ as a new reactive rule to $R_i$, if:
1. *current* $\theta$ is *true* in *Aux* ∪ $S_i^*$ ∪ $ev_i^*$,
2. $\theta$ instantiates all and only the variables in *current*,
3. $\theta$ instantiates all the variable timestamps in FOL conditions in *current* to *i*,
4. *current* contains all the temporal constraints in the antecedent of *r* that
become true in *Aux* as the result of evaluating the FOL conditions in *current*.

(b) If *later* $\theta$ is empty (equivalent to *true*) then delete *later* $\theta$ → *consequent* $\theta$ from $R_i$ and start a new thread in $G_i$ with *consequent* $\theta$ at the root. Add each disjunct of *consequent* $\theta$ whose constraints are satisfiable in *Aux* as a child of the root.

**Step 2. Evaluate goal clauses.** Choose a set of sequencings *current* $\theta$ < *later* $\theta$ of instances $C\theta$ of goal clauses *C*, where *current* is a non-empty complex, from one or more threads in $G_i$. For each such choice, add *later* $\theta$ to $G_i$ as a child of *C*, if:
1. *current* $\theta$ is *true* in *Aux* ∪ $S_i^*$ ∪ $ev_i^*$,
2. $\theta$ instantiates all and only the variables in *current*,
3. $\theta$ instantiates all the variable timestamps in FOL condition*s* in *current* to *i*,
4. *current* contains all the temporal constraints in *C* that become true in *Aux* as
the result of evaluating the FOL conditions in *current*.

**Step 3. Choose candidate actions.** Choose a set of sequencings *actions* $\tau \leq$ *rest* $\tau$ of instances $C \tau$ of goal clauses *C* from one or more threads in $G_i$, where $\tau$ instantiates all and only the timestamp variables in *actions* to *i+1*, and *actions* $\tau$ is the conjunction of *all* the ground bare action atoms in $C \tau$ that have the timestamp *i+1*. Let *candidate-acts*$_{i+1}$ be the set of all the action atoms in all such *actions* $\tau$.

**Step 4. Update** $S_i$, $G_i$, $R_i$. Choose[5] a subset $acts_{i+1}^* \subseteq$ *candidate-acts*$_{i+1}$ such that $C_{pre}$ is *true* in *Aux* ∪ $S_i^*$ ∪ $ev_i^*$ ∪ $ev_{i+1}^*$, where $ev_{i+1}^* = ext_{i+1}^* \cup acts_{i+1}^*$ and the external events $ext_{i+1}^*$ are given. Let $S_{i+1} = succ(S_i, ev_{i+1})$, $G_{i+1} = G_i$ and $R_{i+1} = R_i$.

---

[5] Note that if an action *act* in a goal clause *C* is selected in step 3 and is successfully executed in step 4, then *C* is a candidate to be selected in step 2 of the next cycle, and *act* is a candidate to be selected for evaluation to *true* in *C*. Moreover, other occurrences of *act* in other goal clauses can be evaluated to true in the same cycle, even if they were not selected in step 3 of the previous cycle.

Note that the OS can attempt to make an instance of a consequent of a reactive rule true even though the same instance of the consequent has already been made true. This can be avoided easily in the OS, by adding an extra case (b) to step 2, analogous to case (b) of step 1, but would make the corresponding definition of reactive interpretations more complex. However, there are other optimisations that can be made to the OS, without affecting the definition of reactive interpretation. These optimisations include removing from $R_i$ rules whose antecedents are timed out, and removing from $G_i$ goal clauses containing a fluent or event atom that is timed out.

## 5 Relationships between the Model-theoretic and the Operational Semantics

The proof of soundness for the OS of LPS (Kowalski and Sadri 2010, 2011, 2014), also applies to KELPS, and details of the proof are given in the appendix:

***Theorem 2*. Soundness.** Given a KELPS framework <***R***, ***Aux***, ***C***>, initial state $S_0$ and sequence $ext_1,\ldots, ext_i,\ldots$ of sets of external events, suppose that the OS generates the sequences of sets $acts_1,\ldots, acts_i,\ldots$ of actions and $S_1,\ldots, S_i,\ldots$ of states. Then ***R*** ∪ $C_{pre}$ is true in ***I*** = ***Aux*** ∪ ***S*** * ∪ ***ev*** * if, for every goal tree that is added to a goal state $G_i$, $i \geq 0$, the goal clause *true* is added to the same goal tree in some goal state $G_j$, $j \geq i$.

The following theorem characterises the interpretations generated by the OS. It is a correctness result for reactive interpretations. The detailed proof is in the appendix.

***Theorem 3*. Every interpretation generated by the OS is reactive.** Given a range restricted KELPS framework <***R***, ***Aux***, ***C***>, initial state $S_0$ and set of external events ***ext***\*, let ***acts***\* be the set of actions generated by the OS, and ***ev***\* = ***ext***\* ∪ ***acts***\*. Then ***I*** = ***Aux*** ∪ ***S***\* ∪ ***ev***\* is a reactive interpretation.

**Proof Sketch:** Here is a sketch of the proof for the case where all variables in $X$ and $Y$ in rules $\forall X\,[antecedent \rightarrow \exists Y\,[consequent]]$ have been replaced by all their ground instances, and all the resulting ground temporal constraints have been evaluated, so that the resulting rules no longer contain any temporal constraints:

To show that any action generated by the OS is supported, we show more generally that, if a goal clause $C$ is in a goal state $G_i$, where $i \geq 0$, then there exists a rule in ***R*** of the form *antecedent* → [*other* ∨ [*earlier* ∧ $C$]] such that *earlier* < $C$ and *antecedent* and *earlier* are true in ***Aux*** ∪ $S_0$\* ∪ ... ∪ $S_i$\* ∪ $ev_0$\* ... ∪ $ev_i$\*.

If an action *act* is generated by the OS, then it belongs to some $acts_{i+1}$, $i \geq 0$, and *act* is selected as a candidate in step 3 at time $i$ from a goal clause in $G_i$ that has a sequencing *act* ≤ *rest*. It follows that *r* supports *act*, and ***I*** is supported.

The following theorem is a kind of completeness result for reactive interpretations. The detailed proof is in the appendix.

***Theorem 4*. Every reactive interpretation can be generated by the OS.** Given a range restricted KELPS framework <***R***, ***Aux***, ***C***>, initial state $S_0$ and external events ***ext***\*, let ***acts***\* be a set of actions such that ***I*** = ***Aux*** ∪ ***S***\* ∪ ***ev***\*, where ***ev***\* = ***ext***\*

∪ *acts\**, is a reactive interpretation. Then there exist choices in steps 2, 3 and 4 such that the OS generates *acts\** (and therefore generates *I*).

**Proof Sketch:** Here is a sketch for the case where all the variables *X* and *Y* in rules ∀*X* [*antecedent*→ ∃*Y* [*consequent*]] are replaced by their ground instances, and all temporal constraints have been evaluated:

Let $act_i \in$ ***acts\**** be an action at time *i* supported by a rule $r \in$ ***R***. Then *r* has the form *antecedent* → [*other* ∨ [*early* ∧ $act_i$ ∧ *rest*]], where *antecedent* ∧ *early* is true in ***Aux*** ∪ $S_0^*$ ∪ ... ∪ $S_{i-1}^*$ ∪ $ev_0^*$ ... ∪ $ev_{i-1}^*$. It suffices to show that $G_i$ contains a goal clause of the form $act_i$ ∧ *rest*, because then the OS can choose candidate actions in step 3 and a combination $ev_i^* = ext_i^* \cup acts_i^*$, of external events and candidate actions in step 4, such that $acts_i^*$ is the set of all such actions $act_i$.

To show that $G_i$ contains a goal clause of the form $act_i$ ∧ *rest*, we show more generally that for all times $k \leq i$, and for all $r \in$ ***R*** that support an action in $acts_i^*$ there exist choices in steps 2, 3 and 4 such that either:

- *r* has the form *antecedent*→ [*other* ∨ [*early* ∧ *late*]] where *late* is in $G_k$ and *antecedent* ∧ *early* is true in ***Aux*** ∪ $S_0^*$ ∪ ... ∪ $S_{k-1}^*$ ∪ $ev_0^*$ ...∪ $ev_{k-1}^*$ or
- *r* has the form *earlier*∧ *later* → *consequent*, where *later* → *consequent* is in $R_k$ and *earlier* is true in ***Aux*** ∪ $S_0^*$ ∪ ... ∪ $S_{k-1}^*$ ∪ $ev_0^*$ ... ∪ $ev_{k-1}^*$.

**Frame axioms.** Most logic-based causal theories in AI employ frame axioms to represent and reason about change of state. These frame axioms represent the property that a fluent persists from one state to the next, unless it is terminated by the events that give rise to the state transition. They are used either to reason forwards to copy fluents unchanged from one state to the next, or to reason backwards to determine whether a fluent holds in a state by determining whether holds in the previous state. The use of destructive assignment in LPS and KELPS avoids these computationally expensive forms of reasoning.

In (Kowalski and Sadri 2015), we show that frame axioms are an emergent property of the LPS operational semantics. This is analogous to showing, for example, that associativity of addition is an emergent property of a program that computes addition. The proof for LPS also applies to KELPS. In fact, the proof does not depend on the OS, but only on the definition of $succ(S_i, ev_{i+1})$:

***Theorem 5.*** **Frame axioms are an emergent property.** Given definitions $C_{post}$ of the predicates *initiates* and *terminates*, initial state $S_0$ and sequence of sets of concurrent events $ev_1, ..., ev_i, ...$, let ***I*** = $C_{post}$ ∪ ***S\**** ∪ ***ev\****, where:

  ***S\**** = $S_0^*$ ∪ ... ∪ $S_i^*$ ∪ ... where $S_{i+1} = succ(S_i, ev_{i+1})$ and
  ***ev\**** = $ev_1^*$ ∪ ... ∪ $ev_i^*$ ∪ ....

Then for all $ev_{i+1}$, $i \geq 0$ and fluents *p*, the following sentence is true in ***I***:

  [*initiates*($ev_{i+1}$, *p*) → *p*(*i+1*)] ∧
  [*p*(*i*) ∧ ¬ *terminates*($ev_{i+1}$, *p*) → *p*(*i+1*)]

## 6 Related Work

The development of KELPS/LPS has been influenced by work in many different areas of computing, including programming, databases and artificial intelligence. To make the task of comparison with related work more manageable, we focus primarily on comparing KELPS/LPS with other approaches that attempt to give a logical semantics to rule-based systems. Our earlier papers (Kowalski and Sadri 2009, 2010, 2011, 2012, 2014, 2015) include extensive comparisons between LPS and many of these systems. In this paper, we relate the earlier comparisons with KELPS, and include other related work.

### 6.1 LPS

First, we need to clarify the relationship between KELPS and LPS. Consider, for example, a reactive rule in LPS, which expresses a robot's goal of replying with a sentence to any sentence said by an agent called "*turing*":

*sentence(turing, T1, T2)*→ *sentence(robot, T3, T4)* ∧ *T2 < T3 < T2 + 3 sec*

Here the predicate *sentence* represents a composite event or action with its start and end times. In LPS, the predicate can be defined by a logic program, which can be used both to recognize and to generate sentences. The logic program could include, among other clauses defining the predicates *np* and *vp*, such clauses as:

*sentence(Agent, T1, T3)*← *np(Agent, T1, T2)* ∧ *vp(Agent, T2, T3)*
*adj(Agent, T, T+1)*← *say(Agent, human, T, T+1)*
*noun(Agent, T, T+1)*← *say(Agent, human, T, T+1)*

Here *say* represents a primitive event with its start and end times, In KELPS, such a primitive event is represented more simply with only its end time.

In KELPS, it is possible to get a similar effect to LPS, by replacing the one rule by infinitely many rules, including for example the rule:

*say(turing, i, 2)* ∧ *say(turing, am, 3)* ∧ *say(turing, human, 4)*
→ *say(robot, you, 6)* ∧ *say(robot, are, 7)* ∧ *say(robot, intelligent, 8)*

This is neither practical nor desirable, which is why LPS has both reactive rules and logic programs. As far as we know, other than CHR (section 6.4), there is no other language not derived from ALP that combines reactive rules and logic programs, but retains their separate character. Most languages that have only reactive rules simulate logic programs by treating goals as facts. Most approaches that give a logical semantics to reactive rules translate them into logic programs.

In this paper, we have restricted our attention to reactive rules in KELPS only because it simplifies and clarifies the investigation of completeness.

### 6.2 Abductive Logic Programming (ALP) and the Event Calculus

LPS (and therefore KELPS) originated in our earlier work on ALP (Fung and Kowalski 1997; Kakas et al 1998) and ALP agents (Kowalski and Sadri 1999, 2009; Kowalski 2011). ALP extends logic programming by allowing certain predicates (called *abducible* or *open* predicates) to be assumed in order to solve a problem. These assumptions are restricted by means of integrity constraints. In ALP agents, the abducible predicates represent actions, and the integrity constraints include reactive rules. The main difference is that KELPS/LPS uses destructive updates for state transitions, whereas ALP agents use the event calculus (Kowalski and Sergot 1986). Although the event calculus has been viewed as solving the frame problem (Shanahan 1997), we believe that it cannot compete for practical applications with destructive change of state. However, destructive change of state does not have an obvious logical semantics. In particular, if states are axiomatic theories, which are syntactic objects, then destructive change of state is problematic, because it is not possible to change the axioms during the course of trying to prove a theorem.

KELPS/LPS retains the ontology of the event calculus but replaces explicit reasoning with event calculus axioms by implicit construction of timestamped Herbrand interpretations. As a consequence, KELPS/LPS can generate only reactive models. In contrast, the use of the event calculus in ALP agents can also generate preventative and proactive models.

LPS inherits from ALP the property that models can contain irrelevant actions. In ALP, irrelevant actions can be avoided by minimizing the set of actions contained in a model. In LPS, the generation of irrelevant actions is reduced by generating only reactive interpretations, which contain only actions that are instances of action atoms that occur explicitly in the consequents of reactive rules.

### 6.3 *MetateM*

To the best of our knowledge, MetateM (Barringer *et al* 1996) is the only other framework not based on ALP that describes state transitions by reactive rules in logical form and that treats computation as model generation. Programs in MetateM consist of sentences in modal temporal logic of the form:

'past and present formula' **implies** 'present or future formula'

Computation consists in generating a model in which all such sentences are *true*. MetateM does not formally distinguish between events and fluents, and does not have an explicit causal theory, but frame axioms and other logic programs are written, as needed, in the form of reactive rules.

Like KELPS, MetateM lacks the logic programs of LPS. The main differences are that, in KELPS, time is represented explicitly, models are classical rather than modal, and models are constructed by means of destructive updates. In LPS, there is the further difference that logic programs are separate from and additional to reactive rules.

(Barringer *et al* 1996) presents an operational semantics for propositional MetateM without external events. Despite the claim (on page 148) that the MetateM OS is complete, our examples of non-reactive models for KELPS can be translated into MetateM, and show that the MetateM OS can generate only reactive models, and therefore is incomplete.

### 6.4 Constraint Handling Rules (CHR)

Closer to LPS in syntactic structure and expressive power is the language CHR (constraint handling rules) (Frühwirth 1998, 2009). As in LPS, there are two kinds of rules: equivalences and propagation rules. Equivalences can be used like ordinary logic programs, but can also be used for simplification. Propagation rules are logical implications, which are used to reason forward, like reactive rules in KELPS. $CHR^{\vee}$ (Abdennadher and Schütz 1998) extends CHR to include disjunctions in the consequents of propagation rules, more like reactive rules in KELPS.

The operational semantics of propagation rules in $CHR^{\vee}$ is similar to that of KELPS, and the operational semantics of propagation rules and equivalences is similar to that of LPS. CHR has a classical logic semantics, in which computation is interpreted as theorem-proving, similar to that of the IFF proof-procedure (Fung and Kowalski 1997). This semantics does not justify destructive updates. But CHR also has a translation into linear logic (Betz and Frühwirth, 2005), which justifies destructive updates in the style of LPS, but without the use of timestamps.

The linear logic semantics of CHR is very different from the model-theoretic semantics of KELPS/LPS, in which time is represented explicitly, and state transitions are performed by means of a causal theory. Although propagation rules are similar to reactive rules, it is not clear how the completeness and incompleteness of the KELPS/LPS OS relate to the completeness results for CHR/ $CHR^{\vee}$.

### *6.5 Production systems (PS)*

Production systems are computer languages in which programs are sets of sentences of the form *if conditions then actions*. State transitions are performed by evaluating the *conditions* in the current state and executing the corresponding *actions*, performing "conflict resolution" to choose between conflicting actions. The chosen actions destructively update the current state.

It was the attempt to understand the logic of production systems and their relationship with logic programs that led to the development of LPS (which explains the PS in LPS). Several other authors have also attempted to provide production rules with a logical semantics, mostly by mapping them into logic programs. Raschid (1994), in particular, transforms production rules into logic programs, and uses the fixed point semantics of logic programming to perform forward chaining. Baral and Lobo (1995), on the other hand, translate production rules into the situation calculus represented as a logic program with the stable model semantics.

More recently, Damásio *et al* 2010) use incremental Answer Set Programming (ASP) to realize different conflict resolution strategies for the RIF-PRD production system dialect. Eiter et al. (2012) simulate production systems in ASP with an interface to an external environment, performing state changes by updating and accessing the environment via action atoms and external atoms. Gebser et al. (2011) use disjunctive logic programs (but without existential quantifiers) to represent reactive rules in ASP, for "reasoning about real-time dynamic systems running online in changing environments".

To the best of our knowledge, none of these mappings into logic programs perform destructive updates in the manner of production systems and KELPS/LPS.

In contrast with these approaches, KELPS/LPS reformulates production rules as reactive rules, and not as logic programs. In LPS, logic programs are like "deductive" databases, and reactive rules are like database integrity constraints.

### *6.6 Integrity Constraints*

The model-theoretic semantics of KELPS, which is fundamental to the investigation of completeness in this paper, is based upon the model-theoretic semantics of integrity constraints in deductive databases and ALP. However, from a historical perspective (Kowalski 2014), the model-theoretic semantics is only one of two main alternatives. The other alternative is the theorem-hood view, in which integrity constraints (and queries) are meant to be theorems that are a logical consequence of the database regarded as a theory.

In the parallel world of logic programming, there has been a shift away from the theorem-proving view of computation to a model-generation view. Our own work on the semantics of KELPS/LPS has followed this shift from theorem-proving to model-generation. In LPS, the semantics of logic programs (Kowalski and Sadri 2015) is in the spirit of the well-founded semantics (Van Gelder et al, 1991). But in the case of KELPS, the role of logic programs is played more simply by sets of ground atoms, which are equivalent to relational databases.

Reactive rules in KLEPS/LPS are similar in syntax to integrity constraints in the form of disjunctive tuple-generating dependencies in relational databases (Wang et al. 2001). But, whereas in relational databases the main focus has been on the problem of deciding whether a given set of tuple-generating dependencies logically implies another dependency, in KELPS/LPS the focus is on generating a model that makes integrity constraints in the form of reactive rules $R$ and preconditions $C_{pre}$ true.

In recent years, most of the activity in deductive databases has become associated with Datalog, in which databases are logic programs without function symbols. Datalog± (Calì et al. 2009) extends Datalog with existential rules, which are similar in syntax to reactive rules in KELPS/LPS (but without disjunctive consequents), and which can similarly be viewed as integrity constraints. However, whereas reactive rules and logic programs are separate components of LPS, in Datalog± logic programs are a special case of existential rules. So in effect, Datalog± does not distinguish between the database and integrity constraints. Moreover, it deals only with one database state at a time, and does not deal with database updates.

Datalog± generates answers to queries by using a bottom-up procedure, called the *chase* (Maier *et al*, 1979), which is also used in (Wang et al. 2001). Gavanelli et al (2015) map Datalog± into ALP, and show how the SCIFF proof procedure Alberti et al. 2008), which extends the IFF proof procedure of (Fung and Kowalski, 1997) generates the same answers as the chase procedure. The IFF proof procedure, in turn, is one of the progenitors of the KELPS/LPS OS.

The IFF proof procedure, in turn, is similar to the SATCHMO proof procedure (Manthey and Bry, 1988) for the full clausal form of first-order logic. SATCHMO is a bottom-up procedure, which can be viewed both as a resolution-based theorem-prover and as a model-generator.

### *6.7   Transaction Logic*

Transaction Logic (*TR*) (Bonner and Kifer 1993) is a logic-based formalism for defining transactions, which are similar to complex plans in KELPS/LPS, and which similarly update states of a logic program or database. Transactions in *TR* have a logical, model-theoretic semantics, which, like the possible world semantics of modal logic, is based on sets of possible worlds (or states). But unlike modal logic, the truth value of a transaction is defined along a path of states, starting with the state at the beginning of the transaction and ending with the state at the end of the transaction. As in KELPS/LPS, state transitions are performed by means of destructive updates. Although there is no direct analogue of the reactive rules of KELPS/LPS, they can be simulated by transactions. Rezk and Kifer (2012) use such a simulation to give a logic-based semantics to a production system language.

KELPS shares with *TR* the view of computation as making a goal true by generating a sequence of destructively updated states. Moreover, the inclusion of FOL conditions in KELPS/LPS was largely inspired by similar FOL conditions in *TR* transactions. The main differences are that, in KELPS, transactions are the consequents of reactive rules that are triggered when the antecedents become *true*, time is represented explicitly, and all states, actions and events are combined into a single model-theoretic structure.

$TR^{ev}$ (Gomes and Alferes 2014) extends *TR* by combining the generation of complex transactions with the recognition of complex events. As a consequence, its expressive power is similar to KELPS, but it represents complex events and complex transactions separately, without combining them into reactive rules that are logical implications.

### 6.8 Active Databases

A number of researchers have attempted to develop logic-based semantics for active databases. The majority of these approaches map reactive rules in the form of event-condition-action (ECA) rules into logic programs. For example, both Zaniolo (1993) and Lausen et al. (1998) use frame axioms similar to those in Theorem 5, and reduce ECA rules to logic programs such as $action(T+1) \leftarrow condition(T) \land event(T+1)$. Fernandes et al. (1997) give separate logical formalisations of events, conditions and actions, but without combining them into reactive rules that are logical implications.

Like production systems, and unlike KELPS/LPS, active databases are restricted to rules whose antecedents query the current state, and whose consequents update the current state.

### 6.9 Agent Languages

Bailey et al. (1995) argue that, although they differ in their intended applications and research communities, many agent languages developed in AI are similar to active database systems. In particular, agent plans in BDI agents are similar to ECA rules in active databases. Moreover, both BDI agents and active databases maintain a destructively updated database state, and lack a declarative semantics.

LPS is a direct descendant of ALP agents, which embed ALP in the thinking component of a BDI-like agent cycle. In ALP agents, a logic program represents the agent's beliefs, and initial goals and integrity constraints represent the agent's goals (or desires). The database is updated by using the event calculus, using frame axioms. The ALP agent approach was developed further in the KGP agent model (Mancarella et al. 2009). In contrast, the operational semantics of LPS employs a destructively updated database, which represents the current state.

The destructive updates of LPS were inspired in part by their use in BDI-agent languages such as AgentSpeak (Rao 1996). Programs in AgentSpeak are collections of statements of the form:

*event E: conditions C* $\Leftarrow$ *goals G and actions A.*

The event *E* can be the addition or deletion of a belief literal or a goal atom, stored in a database. The conditions *C* query the database, and the goals *G* and actions *A* update the database by adding or deleting goals and beliefs. As a result, plans combine some of the functionality of both reactive rules and logic programs in LPS. However, they do not allow complex events in the event or conditions part of plans, and they do not include temporal constraints. Moreover, they do not a have a logical semantics. In fact, when *E* is the addition of a belief literal, the arrow $\Leftarrow$ is opposite to the arrow of logical implication in KELPS/LPS.

A number of authors have also developed agent languages and systems in a logic programming context. For example in DALI (Costantini and Tocchio 2014) and EVOLP (Brogi 2002) events transform an initial agent logic program into a sequence of logic programs. ERA (Alferes et al. 2006) extends EVOLP with complex events, complex actions, and event-condition-action rules. The semantics of the evolutionary sequence of logic programs in DALI, EVOLP and ERA is given by an associated sequence of models. In ERA, ECA rules are translated into logic programs of the form *action* $\leftarrow$ *condition* $\wedge$ *event*. In contrast, KELPS/LPS distinguishes between the semantics of logic programs and reactive rules, and combines the sequence of states and events into a single model using timestamps.

FLUX (Thielscher 2005) is a constraint logic programming language for implementing intelligent agents using the fluent calculus. One of the objectives of the fluent calculus is to avoid the computational inefficiency of reasoning with frame axioms, by reifying states as lists of fluents. However, the list representation requires the explicit use of recursion both to query whether a fluent is a member of a list representing a state, and to delete a fluent if it is terminated by an action. Arguably, this is nearly as inefficient as reasoning with explicit frame axioms. In contrast, in KELPS/LPS, states are not represented explicitly in the language, but are represented implicitly by sets of fluents, and membership and deletion are performed by associative look-up. In FLUX, states can be updated by sensing actions, but there seems to be no analogue of the reactive rules of KELPS/LPS.

In (Governatori and Rotolo, 2013), the authors present an efficient algorithm for computing argumentation-style extensions in temporal defeasible logic (TDL). Programs in TDL are logical implications with time stamps, similar to reactive rules in KELPS. The time complexity of the algorithm is proportional to the size of the rules, which is much more efficient than explicit reasoning with event calculus style frame axioms. However, because the algorithm is not described in terms of destructive updates, it is difficult to compare it directly with the OS of KELPS.

Most of these approaches focus on the internal reasoning of individual agents, treating other agents as part of the external environment. In contrast, SCIFF (Alberti et al. 2008) focuses directly on the specification and verification of multi-agent systems, abstracting away from the internal structure of individual agents. It uses ALP to represent the expected behavior of agents and a variant of the IFF proof procedure (Fung and Kowalski 1997), to verify that actual behaviour conforms to expectations.

Integrity constraints in SCIFF resemble ECA rules, but are used to specify the occurrence of events, and not to generate them. Events are timestamped, but are not related to fluents by a causal theory.

### 6.10 Reactive Systems Programming Languages

Although LPS has its origins in logic programming, which is Turing complete, the kernel of LPS is its reactive rules. This shift from logic programs to reactive rules reflects our better appreciation of the fundamental role of state transition systems in all areas of computing, including reactive systems programming languages.

As David Harel (1987) puts it: "Much of the literature also seems to be in agreement that states and events are *a piori* a rather natural medium for describing the dynamic behaviour of a complex system.... A basic fragment of such a description is a *state transition,* which takes the general form *when event Y occurs in state A, if condition C is true at the time, the system transfers to state B*". Harel contrasts such reactive systems with "transformational systems", which transform inputs into outputs in a purely declarative manner. Arguably, LPS reconciles Harel's two kinds of computational systems, with reactive rules providing the reactive part, and logic programs providing structure for the "transformational" part.

In contrast, Shapiro (1989) argues that concurrent logic programming languages are well suited for specifying reactive systems. In these languages, the state of a computation consists of a goal, which is a sequence of atoms and an assignment of values to variables in the goals. Programs are guarded Horn clauses, which have the form *head* ← *guard | body.* Goal atoms that match the *head* and satisfy the *guard* are reduced to the goal atoms in the *body*.

Each goal atom is viewed as a process, and the goal as a whole is viewed as a network of concurrent processes. Processes communicate by instantiating shared logical variables. The external environment is represented by a process whose behaviour is specified by another concurrent logic program.

This approach to reactive systems is very different from that of KELPS/LPS, in which states are sets of time-stamped atoms. Operationally, KELPS/LPS is closer to coordination languages, such as Linda (Carriero and Gelernter, 1989), in which processes interact through the medium of a shared state.

Whereas Harel sees the need for both transformational and reactive systems, and Shapiro reduces both kinds of systems to logic programs, Gurevich (2000) models all varieties of computation by abstract state machines (ASM), programmed by rules of the form *if guard then assignments*. States are abstract, model-theoretic structures consisting of objects and functions. State transitions are performed by evaluating the all the *guards* of rules *if guard then assignments* that are true in the current state and executing all of the corresponding *assignments* destructively and in parallel.

Rules in ASM are similar to condition-action rules in production systems and to reactive rules in KELPS. However, ASM rules are more restricted than KELP/LPS rules, whose antecedents and consequents can involve an entire complex of temporally constrained FOL conditions. Compared with LPS, in which all states are combined in a single model, only the individual states in ASM are model-theoretic in character. Moreover, the *if-then* syntax of guarded assignments in ASM does not mean logical implication.

## 7 Conclusions and Future Work

This paper makes a contribution to analyzing the relationship between the operational and model-theoretic semantics of KELPS. In the future, it would be useful to extend the results to LPS. It would also be interesting to extend the OS to capture more of the non-reactive interpretations that satisfy the model-theoretic semantics.

On the practical side, it would be useful to extend reactive rules to allow more complex event conditions. This extension would not affect the model-theoretic semantics, and can be implemented, for example, by storing a window of past events. It is also important to explore the treatment of concurrency in greater depth, especially in the context of multi-agent systems, in which the external events of a single agent include actions generated by other agents.

There are a number of implementations of LPS. Making some of these available for wider use is a major priority for future work.

### Acknowledgements

We are grateful to the anonymous referees for their careful reading of our earlier submission, and for their many helpful comments.

# Appendix

**Theorem 2.** Given a KELPS framework <***R***, ***Aux***, ***C***>, initial state $S_0$ and sequence $ext_1,\ldots, ext_i,\ldots$ of sets of external events, suppose that the OS generates the sequences of sets $acts_1,\ldots, acts_i,\ldots$ of actions and $S_1,\ldots, S_i,\ldots$ of states. Then ***R*** $\cup$ $C_{pre}$ is true in ***I*** = ***Aux*** $\cup$ ***S*** * $\cup$ ***ev*** * if, for every goal tree that is added to a goal state $G_i$, $i \geq 0$, the goal clause *true* is added to the same goal tree in some goal state $G_j, j \geq i$.

**Proof.** To show $C_{pre}$ is true in ***Aux*** $\cup$ $S^* \cup ev^*$, it suffices to show $C_{pre}$ is true in each ***Aux*** $\cup$ $S_i^* \cup ev_i^* \cup ev_{i+1}^*$. But this is ensured by step 4 of the OS.

To show ***R*** is true in ***Aux*** ∪ *S\** ∪ *ev\**, we need to show that for every rule of the form ∀*X* [*antecedent* → ∃*Y consequent*] in ***R***, whenever some instance *antecedent* σ of the *antecedent* is true in ***I*** then the corresponding instance *consequent* σ of the *consequent* is also true in ***I***. But if *antecedent* σ is true in ***I***, then *antecedent* σ becomes true at some time *i* in ***Aux*** ∪ $S_0^*$ ∪ ... ∪ $S_i^*$ ∪ $ev_0^*$ ∪ ... ∪ $ev_i^*$, and *consequent* σ is added as the root of a new goal tree to the current goal state $G_i$. Each disjunct *consequent*$_j$ σ whose temporal constraints are satisfiable in ***Aux*** is added as a child of the root node.

Clearly, *consequent*$_j$ σ implies *consequent* σ. So if *true* → *consequent*$_j$ σ is true in ***I***, then *consequent* σ is true in ***I***. The truth of *true* → *consequent*$_j$ σ in ***I*** follows from the more general fact that if a goal clause *C* is added in step 2 as a child of a goal clause *C'*, then *C* → *C'* is true in ***I***.

Therefore the existence of a goal state $G_j$ where $i \leq j$ and *true* is added to the same goal tree as *consequent* σ in $G_j$ implies that *consequent* σ is true at time *j*, and therefore *consequent* σ is true in ***I***.

The proof of Theorem 3 uses Lemma 2, which is proved using Lemma 1:

***Lemma 1.*** For $i \geq 0$, let *r* be a rule in $R_i$. Then there exists a rule in *R* of the form *ear* ∧ *con* → *consequent* and a substitution σ that grounds all and only the variables in *ear* such that:
   *ear* σ is *true* in *Aux* ∪ $S_0^*$ ∪ ... ∪ $S_i^*$ ∪ $ev_0^*$ ... ∪ $ev_i^*$
   *ear* σ < *con* σ
   *con* σ → *consequent* σ is *r*.

***Proof.*** Let *n* be the number of applications of step 1 in the derivation of *r*. The proof is by induction on *n*.

**Base case** *n* = *0*: Because *r* was derived by 0 applications of step 1, it follows that *r* ∈ *R*. Then *r* has the form *ear* ∧ *con* → *consequent*, where *ear* is empty (equivalent to *true*). Let σ be the empty substitution. Then:
   *ear* σ is *true* in *Aux* ∪ $S_0^*$ ∪ ... ∪ $S_i^*$ ∪ $ev_0^*$ ... ∪ $ev_i^*$
   *ear* σ < *con* σ
   *con* σ → *consequent* σ is *r*. This proves the base case.

**Inductive step** *n* > *0*: Let *r* be added to some $R_k$ by an application of step 1 of the OS to some rule *r'* in $R_k$, where $k \leq i$. By step 1 of the OS:
  *r'* has the form *current* ∧ *later* → *consequent*, where *current* θ < *later* θ
  *r* has the form *later* θ → *consequent* θ
  *current* θ is *true* in ***Aux*** ∪ $S_k^*$ ∪ $ev_k^*$
  θ instantiates all and only the variables in *current*, and
  θ instantiates all the timestamp variables in FOL conditions in *current* to *k*.
By the inductive hypothesis applied to *r'*, there exists a rule *r\** in ***R*** of the form
  *earlier* ∧ *curr* ∧ *rest* → *conseq* and a substitution σ
  that grounds all and only the variables in *earlier* such that:
  *earlier* σ is *true* in *Aux* ∪ $S_0^*$ ∪ ... ∪ $S_k^*$ ∪ $ev_0^*$ ... ∪ $ev_k^*$
  *earlier* σ < *curr* σ ∧ *rest* σ

*current* is *curr* σ  and *later* is *rest* σ.
Then:
   *earlier* σ θ  ∧  *curr* σ θ  is *true* in Aux ∪ $S_0$* ∪ ... ∪ $S_i$* ∪ $ev_0$* ... ∪ $ev_i$*
   *earlier* σ  θ ∧ *curr* σ θ  < *rest* σ θ
   *rest* σ θ → *conseq* σ θ is *r*. This proves the inductive step.

***Lemma 2***. For $i ≥ 0$, let *C* be a goal clause in $G_i$. Then there exists a rule *r* in *R* of the form *antecedent* → [*other* ∨ [*earlier* ∧ *conds*]] and a substitution σ that grounds all and only the variables in *antecedent* ∧ *earlier* such that:
   *antecedent* σ  ∧ *earlier* σ is *true* in Aux ∪ $S_0$* ∪ ... ∪ $S_i$* ∪ $ev_0$* ... ∪ $ev_i$*
   *earlier* σ  < *conds* σ and
   *conds* σ  is *C*.

***Proof.*** Let *n* be the number of applications of step 2 in the derivation of *C*. The proof is by induction on *n*, and is similar to that of Lemma 1.

**Base case** *n = 0*: If *C* is in $G_0$, then, by the definition of $G_0$, there exists a rule *r* of the form *true* → [*other* ∨ [*earlier* ∧ *C*]] where *earlier* is empty, and *r* has the form required in the statement of the Lemma. If *C* is added in step 1 of the OS to $G_k$, $k ≤ i$, then $R_k$ contains a rule *r* of the form *true* → [*other* ∨ *C*] where *other* ∨ *C* is a new root node added to $G_k$. As a consequence of Lemma 1, there exists a rule in ***R*** of the form *ear* ∧ *con* → *consequent* and a substitution σ that grounds all and only the variables in *ear* such that:
   *ear* σ is *true* in ***Aux*** ∪ $S_0$* ∪ ... ∪ $S_k$* ∪ $ev_0$* ... ∪ $ev_k$*
   *ear* σ  < *con* σ
   *con* σ → *consequent* σ is *r*. So
   *con* σ  is *true*, and *consequent* σ is *other* ∨ *C*.
Let *consequent* have the form [*alternatives* ∨ [*earlier* ∧ *conds*]] where *earlier* is *true* and *conds* σ is *C*. Then σ grounds all and only the variables in *ear* ∧ *con* ∧ *earlier* and:
   *ear* σ ∧ *con* σ  ∧  *earlier* σ is *true* in ***Aux*** ∪ $S_0$* ∪ ... ∪ $S_i$* ∪ $ev_0$* ... ∪ $ev_i$*
   *earlier* σ  < *conds* σ
   *conds* σ is *C*. This proves the base case.

**Inductive step** *n > 0*:   Let *C* be added in step 2 of the OS to $G_k$ as a child of a goal clause *C'*, where *C'* is in $G_k$, $k ≤ i$. By step 2 of the OS:
   *C'* has the form *current* ∧ *later*, where *current* θ < *later* θ,
   *C* has the form *later* θ,
   *current* θ is *true* in ***Aux*** ∪ $S_k$* ∪ $ev_k$*,
   θ instantiates all and only the variables in *current*, and
   θ instantiates all the timestamp variables in FOL conditions in *current* to *k*.
By the inductive hypothesis applied to *C'*, there exists a rule *r* in *R* of the form
   *antecedent* → [*other* ∨ [*earlier* ∧ *curr* ∧ *rest* ]] and a substitution σ
   that grounds all and only the variables in *antecedent* ∧ *earlier* such that:
   *antecedent* σ  ∧ *earlier* σ is *true* in Aux ∪ $S_0$* ∪ ... ∪ $S_k$* ∪ $ev_0$* ... ∪ $ev_k$*
   *earlier* σ  < *curr* σ ∧ *rest* σ
   *current* is *curr* σ   and *later*  is *rest* σ.

Then:
  *antecedent* σ θ ∧ *earlier* σ θ ∧ *curr* σ θ
  is *true* in $Aux \cup S_0^* \cup ... \cup S_i^* \cup ev_0^* ... \cup ev_i^*$
  *earlier* σ θ ∧ *curr* σ θ < *rest* σ θ
  *rest* σ θ is *C*. This proves the inductive step.

***Theorem 3.*** Given a range restricted KELPS framework <***R***, ***Aux***, ***C***>, initial state $S_0$ and set of external events ***ext****, let ***acts**** be the set of actions generated by the OS, and ***ev**** = ***ext**** ∪ ***acts****. Then ***I*** = ***Aux*** ∪ ***S**** ∪ ***ev**** is a reactive interpretation.

***Proof.*** Assume that, for $i \geq 0$, an action *action* τ is added to *candidate-acts*$_{i+1}$ in step 3 and included in *acts*$_{i+1}$ in step 4 of the OS at time *i*. It follows that there exists a sequencing *action* τ ≤ *rest* τ of an instance of a goal clause *action* ∧ *rest* in $G_i$, where τ instantiates only the timestamp variable in *action* to the time *i+1*.

By Lemma 2 there exists a rule *r* in ***R*** of the form *antecedent* → [*other* ∨ [*earlier* ∧ *conds1* ∧ *conds2*] and a substitution σ that grounds all and only the variables in *antecedent* ∧ *earlier* such that:
  *antecedent* σ ∧ *earlier* σ is *true* in $Aux \cup S_0^* \cup ... \cup S_i^* \cup ev_0^* ... \cup ev_i^*$
  *conds1* σ is *action*
  *conds2* σ is *rest*
  *earlier* σ < *action* ∧ *rest*.
It follows that *r* σ τ *supports action* τ, in the sense that:
  a) *action* τ is *conds1* σ τ
  b) *antecedent* σ τ ∧ *earlier* σ τ < *conds1* σ τ ∧ *conds2* σ τ
  c) *antecedent* σ τ ∧ *earlier* σ τ ∧ *conds1* σ τ is *true* in ***I***.
Moreover, step 4 ensures that $C_{pre}$ is *true* in $Aux \cup S_i^* \cup ev_{i+1}^*$. Therefore $C_{pre}$ is *true* in ***I***. Therefore ***I*** is reactive. End of proof.

***Theorem 4.*** Given a range restricted KELPS framework <***R***, ***Aux***, ***C***>, initial state $S_0$ and external events ***ext****, let ***acts**** be a set of actions such that ***I*** = ***Aux*** ∪ ***S**** ∪ ***ev****, where ***ev**** = ***ext**** ∪ ***acts****, is a reactive interpretation. Then there exist choices in steps 2, 3 and 4 such that the OS generates ***acts**** (and therefore generates ***I***).

***Proof.*** Let $R^I$ = {(*r*, σ, *t*) / *r* σ supports an action $act_t$ at time *t*}. We show by induction on *i* that for all times $i \geq 0$, there exist choices in steps 2, 3 and 4 such that:
1) For all (*r*, σ, *t*) ∈ $R^I$, if $i \leq t$ then, at the beginning of the OS cycle at time *i*, either (a) there exists a reactive rule $r_i \in R_i$ such that:
- *r* has the form *earlier* ∧ *later* → *consequent*
- *earlier* σ is true in $Aux \cup S_0^* \cup ... \cup S_{i-1}^* \cup ev_0^* ... \cup ev_{i-1}^*$
- *later* σ → *consequent* σ is an instance of $r_i$ and
- *earlier* σ < *later* σ

or (b) there exists a goal clause $C_i$ in $G_i$ such that:
- *r* has the form *antecedent* → [*other* ∨ [*early* ∧ *late*]]
- *antecedent* σ ∧ *early* σ is true in $Aux \cup S_0^* \cup ... \cup S_{i-1}^* \cup ev_0^* ... \cup ev_{i-1}^*$
- *late* σ is an instance of $C_i$ and
- *antecedent* σ ∧ *early* σ < *late* σ

2) At the end of the OS cycle at time $i-1$, the OS has chosen in step 4 all and only the actions in $acts_i*$. Clearly, (2) implies the statement of the Theorem.

Let $i = 0$ and $(r, \sigma, t) \in \mathbf{R^I}$. If $r$ has the form $true \rightarrow [other \vee [earlier \wedge act \wedge rest]]$, where $r\, \sigma$ supports $act\, \sigma$, then $early \wedge earlier \wedge act \wedge rest$, where $early$ is empty (i.e. $true$), is the desired goal clause $C_0$ in $G_0$. Otherwise, $r$ has the form $later \rightarrow consequent$, where $later$ is not empty, which has the same form as $earlier \wedge later \rightarrow consequent$, where $earlier$ is empty. This is the desired reactive rule $r_0 \in R_0$. So case (1a) holds. (2) also holds, because there are no actions before time 1.

Let $i > 0$ and assume that (1) holds (at the beginning of the cycle at time $i-1$) and that (2) holds (at the end of cycle at time $i-2$). To show that (1) holds at time $i$, let $(r, \sigma, t) \in \mathbf{R^I}$ where $i \leq t$. By the induction hypothesis, either (1a) or (1b) holds for $(r, \sigma, t)$ at time $i-1$. Suppose first that (1a) holds at time $i-1$. Then there exists a reactive rule $r_{i-1} \in R_{i-1}$ such that:
- $r$ has the form $earlier \wedge later \rightarrow consequent$
- $earlier\, \sigma$ is true in $\mathbf{Aux} \cup S_0* \cup ... \cup S_{i-2}* \cup ev_0* ... \cup ev_{i-2}*$
- $later\, \sigma \rightarrow consequent\, \sigma$ is an instance of $r_{i-1}$ and
- $earlier\, \sigma < later\, \sigma$.

If no timestamp in $later\, \sigma$ is equal to $i-1$, then $r_{i-1}$ persists until the end of the cycle, becomes the desired $r_i$ at the beginning of the next cycle, and (1a) holds for $(r, \sigma, t)$ at time $i$. Otherwise, $later$ has the form $current \wedge rest$ where $current\, \sigma$ is true in $\mathbf{Aux} \cup S_{i-1}* \cup ev_{i-1}*$ and $current\, \sigma < rest\, \sigma$. Then step 1 of the OS must evaluate the FOL conditions and temporal constraints in $r_{i-1}$ that have $current\, \sigma$ as an instance, generating a rule $r_i \in R_{i-1}$ such that $rest\, \sigma \rightarrow consequent\, \sigma$ is an instance of $r_i$. Therefore $r_i \in R_{i-1}$ is such that:
- $r$ has the form $earlier \wedge current \wedge rest \rightarrow consequent$
- $earlier\, \sigma \wedge current\, \sigma$ is true in $\mathbf{Aux} \cup S_0* \cup ... \cup S_{i-1}* \cup ev_0* ... \cup ev_{i-1}*$
- $rest\, \sigma \rightarrow consequent\, \sigma$ is an instance of $r_i$ and
- $earlier\, \sigma \wedge current\, \sigma < rest\, \sigma$.

If $rest\, \sigma$ is not empty, then $r_i$ persists until the end of the cycle, becomes the desired $r_i$ at beginning of the next cycle, and (1a) holds for $(r, \sigma, t)$ at time $i$.

If $rest\, \sigma$ is empty, then the OS deletes $r_i$ from $R_i$ and adds a new goal tree to $G_{i-1}$ with root node having $consequent\, \sigma$ as an instance. Because $r\, \sigma$ supports some action $act_t$ at time $t$ where $i-1 \leq t$, then $r$ has the form $antecedent \rightarrow [other \vee [conclusion]]$ where $act_t$ is a bare action conjunction $conclusion$. Then the OS adds to $G_{i-1}$ a goal clause $C$ as a child of the new root node such that:
- $r$ has the form $antecedent \rightarrow [other \vee [conclusion]]$
- $antecedent\, \sigma$ is true in $\mathbf{Aux} \cup S_0* \cup ... \cup S_{i-1}* \cup ev_0* ... \cup ev_{i-1}*$
- $conclusion\, \sigma$ is an instance of $C$ and
- $antecedent\, \sigma \leq conclusion\, \sigma$

If no timestamp in FOL conditions in $conclusion\, \sigma$ is equal to $i-1$, then rewrite $conclusion$ as $early \wedge late$ where $early$ is empty. Then:
- $r$ has the form $antecedent \rightarrow [other \vee [early \wedge late]]$
- $antecedent\, \sigma \wedge early\, \sigma$ is true in $\mathbf{Aux} \cup S_0* \cup ... \cup S_{i-1}* \cup ev_0* ... \cup ev_{i-1}*$
- $late\, \sigma$ is an instance of $C$ and

- *antecedent* σ ∧ *early* σ < *late* σ

$C$ persists until the end of the cycle, becomes the desired $C_i$ at the beginning of the next cycle, and (1b) holds for ($r$, σ, $t$) at time $i$.

Otherwise, *conclusion* has the form *early* ∧ *late* where *early* is not empty, *early* σ is true in $Aux \cup S_{i-1}^* \cup ev_{i-1}^*$, and *early* σ < *late* σ. Let the OS in step 2 choose and evaluate the FOL conditions and temporal constraints in $C$ that have *early* σ as an instance, generating a goal clause $C_i$ in $G_{i-1}$ such that *late* σ is an instance of $C_i$. Then:

- $r$ has the form *antecedent* → [*other* ∨ [*early* ∧ *late*]]
- *antecedent* σ ∧ *early* σ is true in $Aux \cup S_0^* \cup ... \cup S_{i-1}^* \cup ev_0^* ... \cup ev_{i-1}^*$
- *late* σ is an instance of $C_i$ and
- *antecedent* σ ∧ *early* σ < *late* σ

$C_i$ persists until the end of the cycle, becomes the desired $C_i$ at the beginning of the next cycle; and (1b) holds for ($r$, σ, $t$) at time $i$.

Suppose instead that the induction hypothesis holds for (1b). Then there exists a goal clause $C_{i-1}$ in $G_{i-1}$ such that:

- $r$ has the form *antecedent* → [*other* ∨ [*early* ∧ *late*]]
- *antecedent* σ ∧ *early* σ is true in $Aux \cup S_0^* \cup ... \cup S_{i-2}^* \cup ev_0^* ... \cup ev_{i-2}^*$
- *late* σ is an instance of $C_{i-1}$ and
- *antecedent* σ ∧ *early* σ < *late* σ.

If no timestamp in FOL conditions in *late* σ is equal to $i$-1, then $C_{i-1}$ persists until the end of the cycle, becomes the desired $C_i$ at the beginning of the next cycle; and (1b) holds for ($r$, σ, $t$) at time $i$.

Otherwise *late* has the form *current* ∧ *rest* where *current* is not empty, *current* σ is true in $Aux \cup S_{i-1}^* \cup ev_{i-1}^*$, and *current* σ < *rest* σ. Let the OS in step 2 choose and evaluate the FOL conditions and temporal constraints in $C_{i-1}$ that have *current* σ as an instance, generating a goal clause $C_i$ in $G_{i-1}$ such that *rest* σ is an instance of $C_i$. Then:

- $r$ has the form *antecedent* → [*other* ∨ [*early* ∧ *current* ∧ *rest*]]
- *antecedent* σ ∧ *early* σ ∧ *current* σ
  is true in $Aux \cup S_0^* \cup ... \cup S_{i-1}^* \cup ev_0^* ... \cup ev_{i-1}^*$
- *rest* σ is an instance of $C_i$ and
- *antecedent* σ ∧ *early* σ ∧ *current* σ < *rest* σ.

$C_i$ persists until the end of the cycle, becomes the desired $C_i$ at the beginning of the next cycle; and (1b) holds for ($r$, σ, $t$) at time $i$.

To show that (2) holds at time $i$, we need to ensure that steps 3 and 4 of the OS can choose $act_i$ if ($r$, σ, $i$) ∈ $R^I$. But this follows from (1b), which ensures that if $r$ has the form *antecedent* → [*other* ∨ [*earlier* ∧ *action* ∧ *rest*]] where *action* σ = $act_i$ and $r$ σ supports $act_i$, then there exists a goal clause $C_{i-1}$ in $G_{i-1}$ such that *action* σ ∧ *rest* σ is an instance of $C_{i-1}$. It is easy to see that step 3 can include $act_i$ in *candidate-acts$_i$*. Because $C_{pre}$ is true in $I$, step 4 of the OS can choose $act_i$ among the actions generated at the end of the cycle. Moreover, for any other bare action atom *act* in a goal clause in $G_{i-1}$ (whether $i ≤ t$ or $i > t$ for all ($r$, σ, $t$) ∈ $R^I$), whether or not step 3 chooses *act*, step 4 should not choose *act*; and this is possible because $I$ satisfies $C_{pre}$.